# The TUM LapChole dataset for the M2CAI 2016 workflow challenge

Ralf Stauder[1], Daniel Ostler[1,2], Michael Kranzfelder[2,3], Sebastian Koller[2], Hubertus Feußner[2,3], Nassir Navab[1,4]

[1] Computer Aided Medical Procedures, Technische Universität München, Germany
[2] Minimally-invasive Interdisciplinary Therapeutical Intervention, Klinikum rechts der Isar, Technische Universität München, Germany
[3] Department of Surgery, Klinikum rechts der Isar, Technische Universität München, Germany
[4] Computer Aided Medical Procedures, Johns Hopkins University, USA
`{ralf.stauder|daniel.ostler}@tum.de`

**Abstract.** In this technical report we present our collected dataset of laparoscopic cholecystectomies (LapChole). Laparoscopic videos of a total of 20 surgeries were recorded and annotated with surgical phase labels, of which 15 were randomly predetermined as training data, while the remaining 5 videos are selected as test data. This dataset was later included as part of the M2CAI 2016 workflow detection challenge during MICCAI 2016 in Athens.

## 1 Introduction

The community on surgical process modeling, or surgical data science as of recently, is growing and the methods presented in the field are becoming more diverse and sophisticated. Among other challenges, especially the detection of surgical gestures, steps, and phases [1] is a popular task, with an increasing number of contributions in the last few years [2–10].

In order to provide a common basis for comparison of approaches, and to establish a first dataset suitable for machine learning approaches, several minimally-invasive surgeries were recorded at two hospitals in Munich and Strasbourg to provide data for the first M2CAI workflow recognition challenge at the MICCAI conference 2016 in Athens. This work will describe the surgical procedure, the acquisition method and the dataset collected in Munich, as well as provide minimal, preliminary phase detection results on only the data described in this article.

## 2 Medical Procedure

Laparoscopic cholecystectomy was chosen as surgery for this dataset. It is a very common procedure with a highly standardized, yet not fully linear workflow that is comparable even across different hospitals and surgical schools.

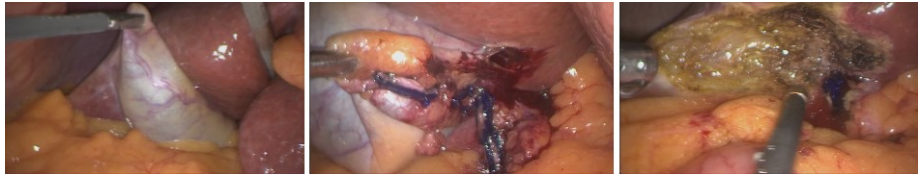

**Fig. 1.** Exemplary images from a recorded surgery. Left: preparation phase. Middle: Clipping and cutting phase. Right: Cleaning and coagulation phase.

With regards to this challenge, the recorded part of the surgery begins with the moment the laparoscopic camera is inserted into the body the first time and finishes with the last time the camera is removed from the body. As we focus on laparoscopic video data for this challenge (see section 3), the patient preparation before this period, and the wound suturing afterwards would not provide any further, meaningful data. We define eight surgical phases for the recorded part of this intervention:

1. Trocar placement
2. Preparation
3. Calot's triangle dissection
4. Clipping and cutting of cystic duct and artery
5. Gallbladder dissection
6. Gallbladder packaging
7. Cleaning and coagulation of liver bed (haemostasis)
8. Gallbladder retraction

The recorded part of the intervention starts with placing the trocars, which will provide access for the laparoscopic instruments. In the following usually short phase the liver is elevated to ease access to the gallbladder and therefore prepared for the surgery. Afterwards Calot's triangle, which is the connection between the gallbladder and the digestive tract, needs to be dissected in order to expose the cystic duct and artery. Both are then sealed by clipping them at least three times each, and cut. After that, the gallbladder needs to be removed from the liver bed and placed in a plastic retrieval bag (gallbladder packaging) to prevent leakage, loss of gallstones, and facilitate extraction from the abdomen. Then the liver bed is checked for bleedings and bile secretion and coagulated. Finally, the gallbladder is removed from the body with the retraction bag.

During any phase it is possible that bleedings occur, which have to be coagulated immediately. It is also possible that a patient has a second, separate cystic artery, which requires an additional clipping and cutting sequence, or that the cystic artery and duct are attached tightly together, which only requires a single clipping and cutting sequence. Moreover, the last phases (6-8) can occur in different order, depending on the performing surgeon and the situational requirements of the intervention.

The recorded dataset starts by definition with phase 1. The transition to the preparation phase is defined by the camera focusing on the liver after the last trocar has been placed. The dissection of the Calot's triangle starts with the surgeon introducing the preparation tool for the first time. When the clipping device is inserted, the fourth phase starts (independently of the type of clip used). When all necessary vessels have been cut and the surgeon reintroduces the dissection tool, phase 5 begins. When the last tissue connecting the gallbladder and the liver is severed, this phase ends and the cleaning and coagulation phase starts. The packaging phase begins when the retraction bag is inserted into the body and ends with the gallbladder fully placed inside. The retraction phase begins with the retraction bag being pulled out of the body, and ends when another instrument is inserted into the body again. Especially phase 7 can occur several times, when the surgeon notices further bleedings after retraction of the gallbladder.

## 3  Data Acquisition

The laparoscopic camera, which is an essential element of every laparoscopic intervention, provides the surgeon with the field of view on the surgical site. Different hardware options are available, including stereoscopic view, though monocular vision with a resolution of 1920x1080 pixels is currently the most common setup. We recorded the view as seen by the surgeon, without and further post-processing. Since the camera system is usually set up significantly before the actual start of the surgery, the raw recording tends to have several minutes of mainly static and uninformative views from the instrument table before inserted into the patient's body. We trimmed the recording to match the start and end of the laparoscopic phases as defined above, and resampled the videos to ensure a constant framerate of 25Hz over all videos.

A total of 20 surgeries were recorded and annotated, of which 15 were randomly selected as training data, the remaining 5 as testing data. For every frame a phase label was assigned, based on the definitions above. Beforehand we obtained approval by the hospital ethics committee to collect the fully anonymized datasets and release them to the public for scientific purposes.

## 4  Phase Detection Baseline

In order to provide a baseline for future comparisons, we applied a readily available machine learning approach to the collected data as described below. Contrary to the final challenge dataset, the following experiment was only conducted on the dataset collected in Munich and described in this paper.

### 4.1  Methodology

We used the standard model of AlexNet [11] and adapted it towards the eight surgical phases as target classes. From the provided video dataset one frame per second was extracted, resulting in 35730 images, from which 25% where used as validation set. Although incidents among classes where unequally distributed, no artificial data augmentation was performed. The model was trained using stochastic gradient descent with a base learning rate of 0.01 and a step down policy with gamma=0.1 with step size of 33% over 30 epochs.

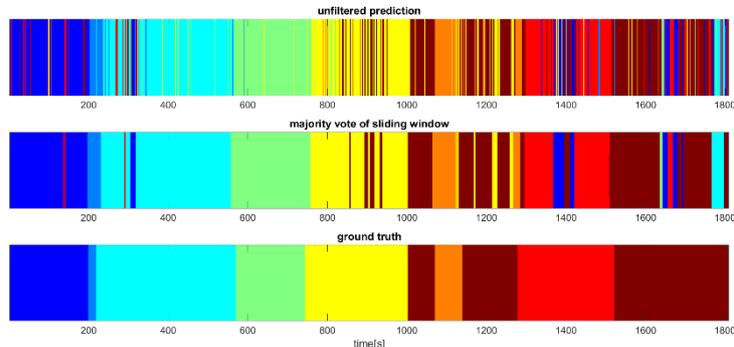

**Fig. 2.** Visualization of classification result on one test surgery. Top: Unfiltered classification output. Middle: Classification output after applying the described sliding window approach. Bottom: Ground truth labels.

For evaluation consecutively one frame per second was extracted from the test videos and classified. In order to reduce the impact of misclassifications on the prediction of the surgical phase, a sliding window approach was applied. Within a sliding window of width 10 seconds (i.e. the last 10 classifications) a majority vote for the current prediction is performed. Every frame is classified with this approach, so the sliding windows for neighboring frames overlap.

### 4.2 Results

We trained the network on the training set described in section 3 and evaluated its performance on the predefined testing set. We achieved an average Jaccard index of 52.4% over all phases and surgeries, with an average precision of 65.9% and an average Recall of 74.7%. The average performance metrics for each phase over all surgeries is given in table 1.

It should be noted, that in one surgery the phase "gallbladder retraction" was not detected at all, so the theoretically undefined precision for that case was set to 0 for further calculations.

## 5  Conclusion

We recorded 20 laparoscopic cholecystectomies and annotated them with surgical phase labels. A first phase detection based on a readily available machine learning framework yields a mean Jaccard index of 52.4% over all phases. These recorded surgeries will be incorporated into the dataset of the M2CAI 2016 workflow detection challenge. With this effort we hope to offer a suitable training dataset for advanced machine learning approaches in surgical data science, and provide a common base for future evaluation and comparison of workflow recognition methods.

| Phase | Precision | Recall | Jaccard |
|---|---|---|---|
| Trocar placement | 57,8% | 89,5% | 53,1% |
| Preparation | 34,1% | 88,8% | 33,8% |
| Calot's triangle | 96,1% | 69,8% | 67,2% |
| Clipping/Cutting | 76,1% | 87,9% | 69,0% |
| Gallbladder dissection | 63,7% | 77,5% | 48,5% |
| Cleaning/Coagulation | 65,8% | 70,4% | 52,7% |
| Gallbladder packaging | 70,5% | 59,6% | 50,4% |
| Gallbladder retraction | 62,6% | 53,9% | 44,3% |

**Table 1.** Performance metrics per phase, averaged over all surgeries

## Acknowledgement

This work was supported by the Bavarian Research Foundation (BFS).